\documentclass[11pt]{article}
\usepackage{coling2020}
\usepackage{times}
\usepackage{url}
\usepackage{latexsym}
\usepackage{graphicx}
\usepackage{amsmath}
\usepackage{amsthm}
\usepackage{booktabs}
\usepackage{algorithm}
\usepackage{algorithmic}
\usepackage{multirow}
\usepackage{subfigure}

\colingfinalcopy 

\title{Regularized Attentive Capsule Network for Overlapped Relation Extraction}

\author{Tianyi Liu$^{1}$\footnotemark[2], Xiangyu Lin$^{2}$\footnotemark[2], Weijia Jia$^{3,1}$\thanks{\ \ Corresponding author: jiawj@sjtu.edu.cn}, Mingliang Zhou$^4$, Wei Zhao$^5$\\
$^1$Department of Computer Science and Engineering, Shanghai Jiao Tong University\\
$^2$Department of Computer and Information Science, University of Macau\\
$^3$BNU-UIC Institute of AI and Future Networks, Beijing Normal University (Zhuhai)\\
$^4$School of Computer Science, Chongqing University\\
$^5$American University of Sharjah, Sharjah United Arab Emirates\\}

\footnotetext[2]{make equal contributions}

\date{}

\begin{document}
\maketitle
\begin{abstract}
  Distantly supervised relation extraction has been widely applied in knowledge base construction due to its less requirement of human efforts. However, the automatically established training datasets in distant supervision contain low-quality instances with noisy words and overlapped relations, introducing great challenges to the accurate extraction of relations. To address this problem, we propose a novel Regularized Attentive Capsule Network (RA-CapNet) to better identify highly overlapped relations in each informal sentence. To discover multiple relation features in an instance, we embed multi-head attention into the capsule network as the low-level capsules, where the subtraction of two entities acts as a new form of relation query to select salient features regardless of their positions. To further discriminate overlapped relation features, we devise disagreement regularization to explicitly encourage the diversity among both multiple attention heads and low-level capsules. Extensive experiments conducted on widely used datasets show that our model achieves significant improvements in relation extraction.
\end{abstract}

\section{Introduction}
%
%
    %
    %
    %
    
    %
    %

Relation extraction aims to extract relations between entities in text, where distant supervision proposed by~\cite{mintz2009distant} automatically establishes training datasets by assigning relation labels to instances that mention entities within knowledge bases. However, the wrong labeling problem can occur and various multi-instance learning methods~\cite{riedel2010modeling,hoffmann2011knowledge,surdeanu2012multi} have been proposed to address it. Despite the wrong labeling problem, each instance in distant supervision is crawled from web pages, which is informal with many noisy words and can express multiple similar relations. This problem is not well-handled by previous approaches and severely hampers the performance of conventional neural relation extractors. To handle this problem, we have to address two challenges: (1) Identifying and gathering spotted relation information from low-quality instances; (2) Distinguishing multiple overlapped relation features from each instance.

\begin{figure}[htbp]
    \centering
    \includegraphics[width=15.5cm]{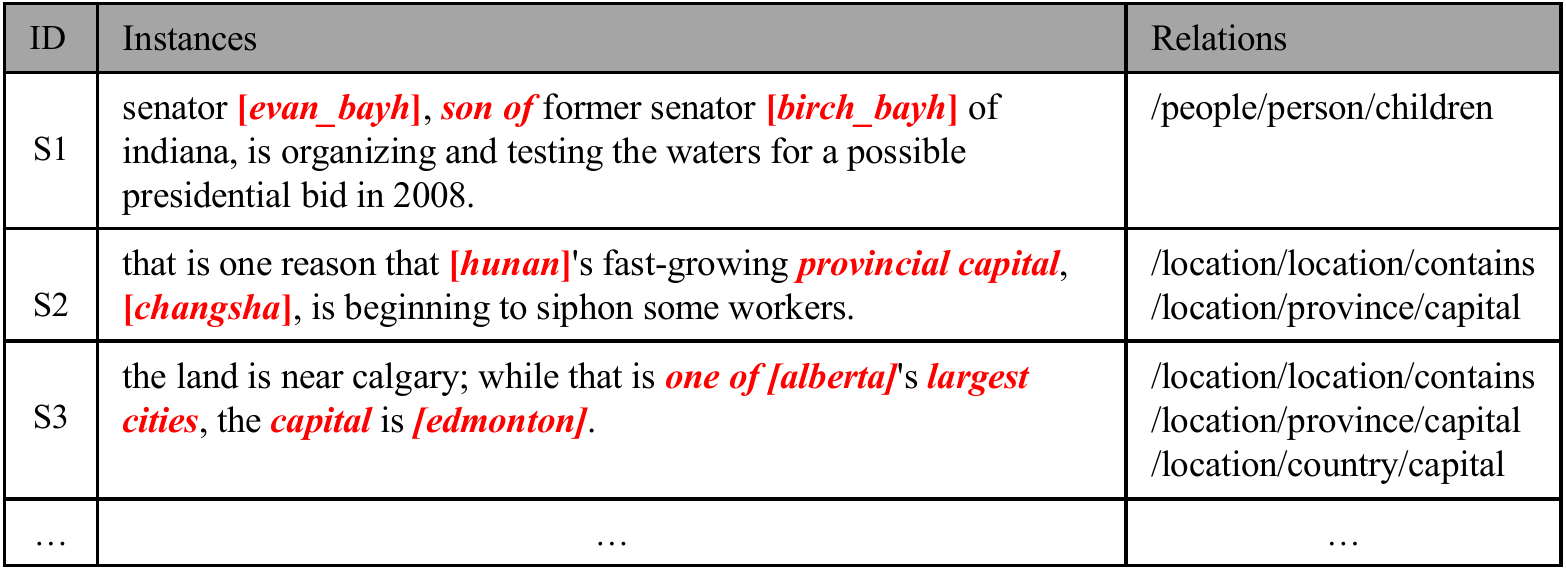}
    \caption{Example of instances from the New York Times (NYT).}
    \label{fig:multi_relation}
 \end{figure}

First, a few significant relation words are distributed dispersedly in the sentence, as shown in Figure~\ref{fig:multi_relation}, where words marked in red brackets represent entities, and italic words are key to expressing the relations. For instance, the clause ``\textit{evan\_bayh son of birch\_bayh}" in \textit{S1} is sufficient to express the relation \textit{/people/person/children} of \textit{evan\_bayh} and \textit{birch\_bayh}. Salient relation words are few in number and dispersedly in \textit{S1}, while others excluded from the clause can be regarded as noise. Traditional neural models have difficulty gathering spotted relation features at different positions along the sequence because they use Convolutional Neural Network (CNN) or Recurrent Neural Network (RNN) as basic relation encoders~\cite{zeng2015distant,liu2018neural,ye2019distant}, which model each sequence word by word and lose rich non-local information for modeling the dependencies of semantic salience. Thus, a well-behaved relation extractor is needed to extract scattered relation features from informal instances.

Second, each instance can express multiple similar relations of two entities. As shown in Figure~\ref{fig:multi_relation}, \textit{Changsha} and \textit{Hunan} possess the relations \textit{/location/location/contains} and \textit{/location/province/capital} in \textit{S2}, which have similar semantics, introducing great challenges for neural extractors in discriminating them clearly. Conventional neural methods are not effective at extracting overlapped relation features, because they mix different relation semantics into a single vector by max-pooling~\cite{zeng2014relation} or self-attention~\cite{lin2016neural}. Although \cite{zhang2019multi} first propose an attentive capsule network for multi-labeled relation extraction, it treats the CNN/RNN as low-level capsules without the diversity encouragement, which poses the difficulty of distinguishing different and overlapped relation features from a single type of semantic capsule. Therefore, a well-behaved relation extractor is needed to discriminate diverse overlapped relation features from different semantic spaces.

To address the above problem, we propose a novel Regularized Attentive Capsule Network (RA-CapNet) to identify highly overlapped relations in the low-quality distant supervision corpus. First, we propose to embed multi-head attention into the capsule network, where attention vectors from each head are encapsulated as a low-level capsule, discovering relation features in an unique semantic space. Then, to improve multi-head attention in extracting spotted relation features, we devise relation query multi-head attention, which selects salient relation words regardless of their positions. This mechanism assigns proper attention scores to salient relation words by calculating the logit similarity of each relation representation and word representation. Furthermore, we apply disagreement regularization to multi-head attention and low-level capsules, which encourages each head or capsule to discriminate different relation features from different semantic spaces. Finally, the dynamic routing algorithm and sliding-margin loss are employed to gather diverse relation features and predict multiple specific relations. We evaluate RA-CapNet using two benchmarks. The experimental results show that our model achieves satisfactory performance over the baselines. Our contributions are summarized as follows: 
\begin{itemize}
    \setlength{\itemsep}{0pt}
    \setlength{\parsep}{0pt}
    \setlength{\parskip}{0pt}
    \item We first propose to embed multi-head attention as low-level capsules into the capsule network for distantly supervised relation extraction.
    \item To improve the ability of multi-head attention in extracting scattered relation features, we design relation query multi-head attention.
    \item To discriminate overlapped relation features, we devise disagreement regularization on multi-head attention and low-level capsules.
    \item RA-CapNet achieves significant improvements for distantly supervised relation extraction.
\end{itemize}

\section{Related Work}

Distantly supervised relation extraction has been essential for knowledge base construction since ~\cite{mintz2009distant} propose it. To address the wrong labeling problem in distant supervision, multi-instance and multi-label approaches are proposed~\cite{riedel2010modeling,hoffmann2011knowledge,surdeanu2012multi}.

With the renaissance of neural networks, increasing researches in distant supervision have been proposed to extract precise relation features. Piecewise CNNs with various attention mechanisms are proposed~\cite{zeng2015distant,lin2016neural,ji2017distant}. Reinforcement learning and adversarial training are proposed to select valid instances to train relation extractors~\cite{feng2018reinforcement,qin2018robust,qin2018dsgan}. Recently, multi-level noise reduction is designed by ~\cite{ye2019distant,jia2019arnor}.

Nevertheless, the above approaches ignore the effect of noisy words and overlapped relation features in each instance. To reduce the impact of noisy words, tree-based methods attempt to obtain the relevant sub-structure of an instance for relation extraction~\cite{xu2015classifying,miwa2016end,liu2018neural}. To discriminate overlapped relation features, ~\cite{zhang2019multi} apply the capsule network~\cite{sabour2017dynamic} for multi-labeled relation extraction. Inspired by the ability of multi-head attention in modeling the long-term dependency~\cite{vaswani2017attention}, ~\cite{8952645} attempt to reduce multi-granularity noise via multi-head attention in relation extraction. 

\begin{figure*}[htbp]
  \centering
  \includegraphics[width=15.5cm]{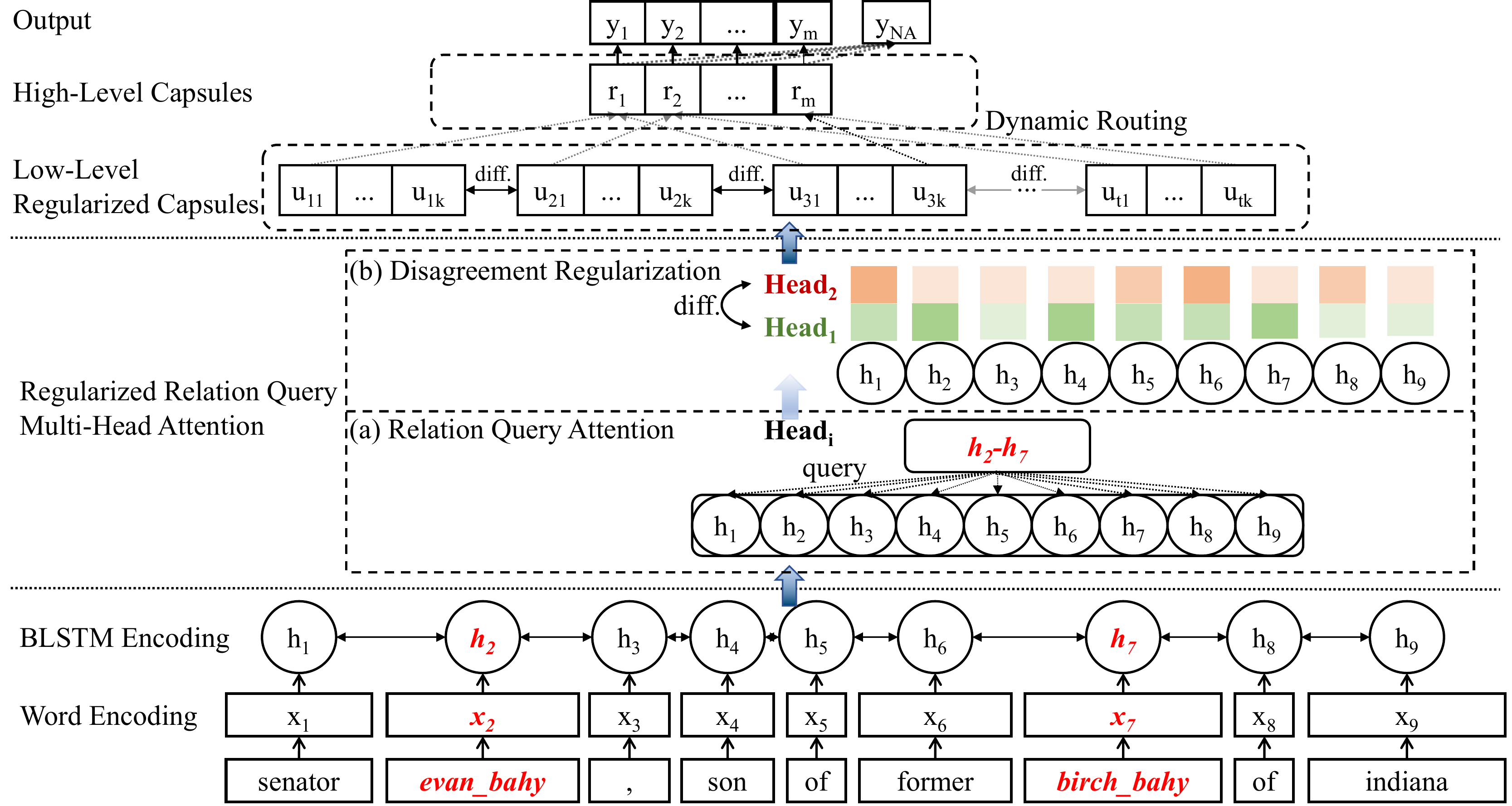}
  \caption{Overall architecture of RA-CapNet, expressing the process of handling an instance.}
  \label{fig:model_framework}
\end{figure*}

\section{Methodology}
As shown in Figure~\ref{fig:model_framework}, we will introduce the three-layer RA-CapNet: (1) \textbf{The Feature Encoding Layer} primarily contains the word encoding layer and BLSTM encoding layer. (2) \textbf{The Feature Extracting Layer} chiefly includes relation query multi-head attention and disagreement regularization. (3) \textbf{The Relation Gathering Layer} mainly consists of a regularized capsule network and dynamic routing.

\subsection{Feature Encoding Layer}
Each instance is first input into the encoding layer to be transformed to the distributed representations for the convenience of calculation and extraction by neural networks.

\subsubsection*{Word Encoding Layer}
As mentioned in ~\cite{zeng2014relation}, the inputs of the relation extractor are word and position tokens, which are encoded by word embeddings and position embeddings at first. Then, the $j_{th}$ input word $x_{ij}$ in the $i_{th}$ instance, is concatenated by one word embedded vector $x_{ij}^w \in R^k$ and two position embedded vectors $x_{ij}^{p1}$ and $x_{ij}^{p2} \in R^p$, $x_{ij}=[x_{ij}^w;x_{ij}^{p1};x_{ij}^{p2}]$, where $k$ and $p$ represent the dimensions of word vectors and position vectors respectively, and $;$ denotes the vertical concatenating operation. To simplify the mathematical expression, we denote $x_{ij}$ as $x_j$.


 
 \subsubsection*{BLSTM Encoding Layer}
 To further encode relation features inside the context, we adopt the Bidirectional Long-Short Term network (BLSTM)~\cite{graves2013generating} as our basic relation encoder, which can access the future context as well as the past. The encoding feature vector $h_i$ of the $i_{th}$ word is calculated as follows:
 \begin{align}
    &\overrightarrow{h_i}=\overrightarrow{LSTM}(x_i,\overrightarrow{h_{i-1}}) \\
    &\overleftarrow{h_i}=\overleftarrow{LSTM}(x_i,\overleftarrow{h_{i+1}}) \\
    &h_i=\overrightarrow{h_i}+\overleftarrow{h_i}
 \end{align}
 where $\overrightarrow{h_i}$ and $\overleftarrow{h_i} \in R^d$ are hidden state vectors of the LSTM. Finally, we obtain the sentence encoding vector $H=[h_1,h_2,\cdots,h_l]$, where $l$ represents the instance length.
 
 \subsection{Feature Extracting Layer}
 First, relation query multi-head attention is devised to emphasize spotted relation features from different semantic spaces. Then, disagreement regularization is applied to encouraging the diversity of relation features that each head discovers.
 
 \subsubsection*{Relation Query Multi-Head Attention}
 
 Multi-head attention is useful for modeling the long-term dependency of salient information in the context~\cite{vaswani2017attention}. Based on this mechanism, we propose relation query multi-head attention to improve the ability of multi-head attention in extracting spotted and salient relation features regardless of their irregular positions in the instance. 
 
 Formally, given an encoding instance $H$, we use the subtraction of two entities' states $h_{en1}$ and $h_{en2}$ as the relation representation, as inspired by ~\cite{bordes2013translating}. The relation representation acts as a query vector as follows: 
 \begin{equation}
     Q^{rel}=(h_{en1}-h_{en2})W^Q \label{eq:q}
 \end{equation}
 where $W^Q \in R^{d \times d}$ is a weight matrix. The corresponding key $K$ and value $V$ vectors are defined:
 \begin{equation}
     K=HW^K \quad V=HW^V \label{eq:kv}
 \end{equation}
 where $W^K$ and $W^V \in R^{d \times d}$ are weight matrices. Afterward, we calculate the logit similarity of the relation query vector and word representation vectors as attention scores:
 \begin{equation}
     energy=\frac{Q^{rel}K^T}{\sqrt{d}}
 \end{equation}
 where the $energy$ can measure the importance of each word to relation extraction, which is leveraged to select salient and spotted relation features along the sequence:
 \begin{equation}
     ATT=softmax(energy)V
 \end{equation}
 To extract diverse relation features, we employ relation query attention into multi-head attention:
 \begin{align}
     &head_i=ATT(Q_i^{rel},K_i,V_i) \label{eq:head} \\
     &E^m=[head_1;head_2;\cdots;head_n]W^o
 \end{align}
 where $W^o \in R^{d \times d}$ is the weight matrix. Multiple heads can capture various semantic features.
 
 After we acquire the output $E^m$ of multi-head attention, a Feed-Forward Network (FFN) is applied:
 \begin{equation}
     H^r=max(0,E^mW^f_1+b^f_1)W^f_2+b^f_2
 \end{equation}
 where $W^f_1 \in R^{d \times d}$, $W^f_2 \in R^{d \times d'}$, $b^f_1 \in R^{d}$ and $b^f_2 \in R^{d'}$ are parameters.
 
\subsubsection*{Disagreement Regularization on Multi-Head Attention}
To further discriminate overlapped relation features from different heads in multi-head attention, we introduce the disagreement regularization based on ~\cite{yang2018modeling}.

Formally, given $n$ heads $Head=[head_1,head_2,\cdots,head_n]$ as calculated in Eq.~(\ref{eq:head}), we calculate the cosine similarity $cos(.)$ between the vector pair $head_i$ and $head_j$ in different value subspaces:
\begin{equation}
    D^{sub}_{ij}=cos(head_i,head_j)=\frac{head_i \cdot head_j}{\|head_i\|\|head_j\|}
\end{equation}
where $\|*\|$ represents the $L2$ norm of vectors. The average cosine distance among all heads is obtained:
\begin{equation}
    D^{sub}=\frac{\sum_{ij}{D^{sub}_{ij}}}{n^2}
\end{equation}
Our goal is to minimize $D^{sub}$, which encourages the heads to be different from each other, improving the diversity of subspaces among multiple heads. Accordingly, each head can discriminate overlapped relation features more clearly.

\subsection{Relation Gathering Layer}
To form relation-specific features, the relation gathering layer gathers scattered relation features from diverse low-level capsules using a dynamic routing algorithm.

\subsubsection*{Low-Level Capsules with Disagreement Regularization}
The capsule network has been proven effective in discriminating overlapped features~\cite{sabour2017dynamic,zhang2019multi}. In our application, a capsule is a group of neural vectors within one-head attention and regularized by a disagreement term. Thus, each capsule can capture relation features in an unique semantic space.  In detail, the orientation of the attention vector inside one head indicates one certain factor of a specific relation, while its length means the probability that this relational factor exists.

We reorganize each attention head of $H^r$ to form a low-level capsule denoted as $u \in R^{d_u}$, where each capsule captures information in a specific semantic space. Formally, the above process is as follows:
\begin{align}
    &H^r=[h^r_1;...;h^r_{t}]\\
    &u_k = g(h^r_k) = \frac{\left \| h^r_k \right \|^{2}}{1+\left \| h^r_k \right \|^{2}}\frac{h^r_k}{\left \| h^r_k \right \|} \label{eq:squash}
\end{align}
where $t$ is the number of low-level capsules, which equals the quantity of heads. Eq.~(\ref{eq:squash}) is a squash function, shrinking the length of vectors from 0 to 1 to express the probability.

To encourage the diversity of these capsules, disagreement regularization is applied to them:
\begin{align}
    &D^{cap}_{ij}=\frac{u_i\cdot u_j}{\|u_i\|\|u_j\|}\\
    &D^{cap}=\frac{\sum_{ij}{D^{cap}_{ij}}}{t^2}
\end{align}
To minimize $D^{cap}$, we can encourage the capsules to be different from each other, improving the diversity of subspaces among multiple capsules and discriminating overlapped relation features more clearly.

The final disagreement regularization term is the average of multi-head and capsule disagreement: 
\begin{equation}
    D = \frac{D^{sub} + D^{cap}}{2} 
\end{equation}
where $D$ is the final disagreement regularization term which only works for the training process.
\subsubsection*{High-Level Capsules with Dynamic Routing}
After the low-level capsules capturing the different aspects of semantic information, the high-level capsules $r \in R^{d_r}$ are produced from them to gather scattered information and form specific relation features, which are calculated as follows:
\begin{equation}
    r_j = g(\sum c_{ij}W^h_j u_i)
\end{equation}
where $W^h_j \in R^{d_u\times d_r}$ are parameters for high-level capsules and $c_{ij}$ are coupling coefficients that are determined by the dynamic routing process described in ~\cite{sabour2017dynamic}.

\subsubsection*{Loss Function}
The sliding-margin loss function used in the capsule network enables the prediction of multiple overlapped relations, which sums up the loss for both the relations present and absent from the instances. This margin loss function is integrated into our model as follows:
\begin{align}
    L_j = & Y_jmax(0,(S+\gamma)-\left \| r_j \right \|)^{2} + \nonumber \\
     & \lambda(1-Y_j)max(0,\left \| r_j \right \|-(S-\gamma))^{2}
\end{align}%
where $\gamma$ is the width of the margin, $S$ is a learnable threshold for ``no relation" (NA), and $\lambda$ is the down-weighting of the loss for absent relations. $Y_j$ = 1 if the relation corresponding to $r_j$ is present in the sentence and $Y_j$ = 0 otherwise.

Afterward, the final loss is defined as follows:
\begin{equation}
    loss=\sum_j{L_j}+\beta D+\beta'\|\theta\|^2
\end{equation}
where $\beta$ and $\beta'$ are hyperparameters used to restrict the disagreement regularization and $L2$ regularization of all parameters $\theta$. In this paper, we use the Adam~\cite{kingma2014adam} to minimize the final loss.

\section{Experiments}
Our experiments are devised to demonstrate that RA-CapNet can identify highly overlapped relations of informal instances in distant supervision. In this section, we first introduce the dataset and experimental setup. Then, we evaluate the overall performance of RA-CapNet and the effects of different parts of RA-CapNet. Finally, we present the case study.

\subsection{Dataset and Experimental Setup}
\textbf{Dataset.} To evaluate the effects of RA-CapNet, we conduct experiments on two datasets. \textbf{NYT-10} is a standard dataset constructed by ~\cite{riedel2010modeling}, which aligns relational tuples in Freebase~\cite{bollacker2008freebase} with the corpus of New York Times. Sentences from 2005-2006 are used as the training set, while sentences from 2007 are used for testing. \textbf{NYT-18} is a larger dataset constructed by ~\cite{8952645} with the same creation method as NYT-10, which crawls 2008-2017 contexts from the NYT. All the sentences are divided into five parts with the same relation distribution for five-fold cross-validation. The details of the datasets are illustrated in Table~\ref{table:dataset}.

  \begin{table}[htbp]
    \centering
    \begin{tabular}{cccccc}
      \toprule
      \multirow{2}{*}{Datasets} & \multicolumn{2}{c}{Training (k)} & \multicolumn{2}{c}{Testing (k)} & \multirow{2}{*}{Rel.} \\
       & Sen. & Ent. & Sen. & Ent. &\\
      \midrule
      NYT-10 & 523 & 281 & 172 & 97 & 53\\
      NYT-18 & 2446 & 1234 & 611 & 394 & 503\\
      \bottomrule
    \end{tabular}
    \caption{The dataset information. \textbf{Sen.}, \textbf{Ent.} and \textbf{Rel.} indicate numbers of sentences, entity pairs and relations (including NA).}
    \label{table:dataset}
  \end{table}

\noindent \textbf{Evaluation Metric.} As mentioned in ~\cite{mintz2009distant}, we use the held-out metrics to evaluate RA-CapNet. The held-out evaluation offers an automatic way to assess models with the PR curve and Precision at top 100 or 10k predictions (P@100 on NYT-10 or P@10k on NYT-18) at all numbers of instances under each entity pair, which indicates that all instances under the entity pair are used to represent the relation.

\noindent \textbf{Parameter.} In our work, we use the \textit{Skip-Gram}~\cite{mikolov2013efficient} to pretrain our word embedding matrices. The words of an entity are concatenated when it has multiple words. The grid search and cross-validation are used to adjust important hyperparameters of the networks. Our final parameter settings are illustrated in Table~\ref{table:parameters}.

\begin{table}[htbp]
    \centering
    \begin{tabular}{ccc}
        \toprule
        Parameter & NYT-10 & NYT-18 \\
        \midrule
        Batch size $b$ & 50 & 50 \\
        Word embedding size $k$ & 50 & 360 \\
        Position embedding size $p$ & 5 & 5 \\
        Sentence length $l$ & 100 & 100 \\
        LSTM hidden size $d$ & 256 & 256 \\
        Multi-head number $n$ & 16 & 16 \\
        FFN hidden size $d'$ & 512 & 512 \\
        Capsule dimensions $[d^u,d^r]$ & [16,16] & [16,16] \\
        Low-level capsule number $t$ & 16 & 16\\
        Valid relation class $m$ & 52 & 502 \\
        Sliding margin $\gamma$ & 0.4 & 0.4 \\
        Down-weighting $\lambda$ & 1.0 & 1.0 \\
        Learning rate $lr$ & 0.0001 & 0.0001 \\
        Dropout probability $p$ & 0.5 & 0.5 \\
        Weight of disagreement $\beta$ & 0.001 & 0.001 \\
        $L2$ penalty $\beta'$ & 1e-08 & 0.0 \\
        \bottomrule
    \end{tabular}
    \caption{Parameter settings.}
    \label{table:parameters}
\end{table}

\subsection{Overall Performance}

\begin{figure}[htbp]
    \centering
    \subfigure{
    \begin{minipage}[t]{7.5cm}
    \centering
    \includegraphics[scale=0.5]{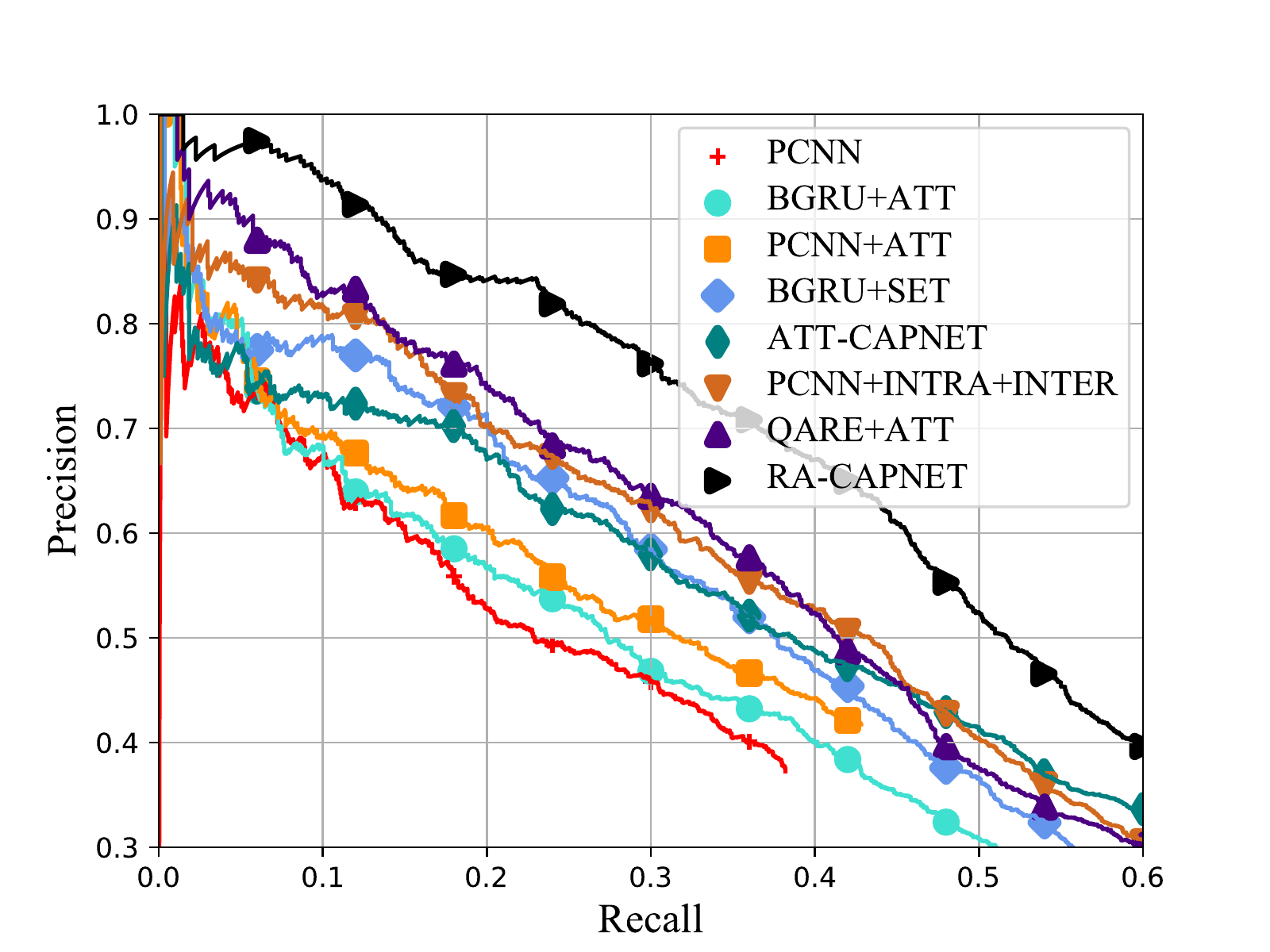}
    \caption{Precision-recall curves on NYT-10.}
    \label{fig:pr-10}
    \end{minipage}
    }
    \subfigure{
    \begin{minipage}[t]{7.5cm}
    \centering
    \includegraphics[scale=0.5]{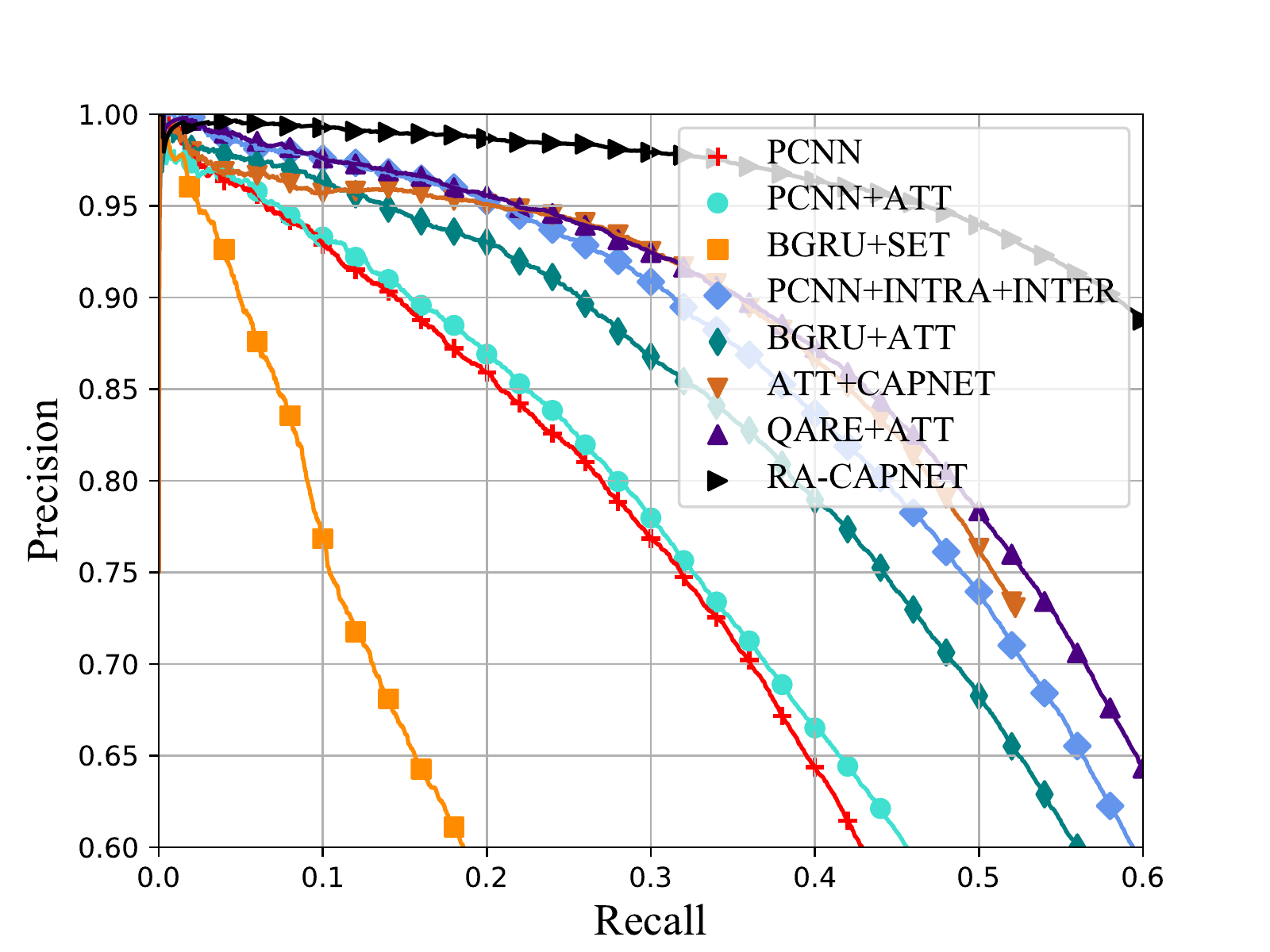}
    \caption{Precision-recall curves on NYT-18.}
    \label{fig:pr-18}
    \end{minipage}
    }
  \end{figure}

To evaluate our model, we select the following methods for comparison:

\textbf{PCNN}~\cite{zeng2015distant} present a piecewise CNN for relation extraction.

\textbf{PCNN+ATT}~\cite{lin2016neural} propose the selective attention mechanism with PCNN.

\textbf{BGRU+ATT}~\cite{zhou2016attention} present a BGRU-based model with word-level attention.

\textbf{BGRU+SET}~\cite{liu2018neural} propose a BGRU-based approach to reduce inner-sentence noise.

\textbf{PCNN+INTRA+INTER }~\cite{ye2019distant} propose to emphasize true labeled sentences and bags.

\textbf{ATT+CAPNET}~\cite{zhang2019multi} put forward an attentive capsule network for relation extraction.

\textbf{QARE+ATT}~\cite{8952645} propose improved multi-head attention with transfer learning.

We compare our method with baselines on two datasets. For both datasets, the PR curves on NYT-10 and NYT-18 are shown in Figure~\ref{fig:pr-10} and Figure~\ref{fig:pr-18}. We find that: (1) BGRU+SET performs well on NYT-10 but poorly on NYT-18. This demonstrates that BGRU+SET is not well-handled on highly informal instances because the complex instances in NYT-18 are difficult to be parsed precisely by the conventional parser. (2) RA-CapNet achieves the best PR curve among all baselines on both datasets, which improves the PR curve significantly. This verifies that our model is effective in capturing overlapped and scattered relation features. (3) RA-CapNet outperforms ATT+CAPNET, which indicates that the relation query multi-head attention and disagreement regularization are useful for overlapped relation extraction.

\begin{table}[htbp]
    \begin{minipage}[b]{7.5cm}
        \centering
        \setlength{\tabcolsep}{0.7mm}{
        \begin{tabular}{lcc}
            \toprule
            \multirow{2}{*}{Model} & \multicolumn{2}{c}{PR curve area} \\
             & NYT-10 & NYT-18 \\
            \midrule
            BGRU+ATT & 0.337 & 0.596 \\
            PCNN+ATT  & 0.356 & 0.511 \\
            BGRU+SET  & 0.392 & 0.290 \\
            ATT-CAPNET  & 0.415 & 0.647 \\
            PCNN+INTRA+INTER  & 0.423 & 0.617 \\
            QARE+ATT & 0.428 & 0.645\\
            RA-CAPNET  & \textbf{0.526} & \textbf{0.780} \\
            \bottomrule
        \end{tabular}}
        \caption{Precision-recall curve areas.}
        \label{table:pr}
    \end{minipage}
    \begin{minipage}[b]{7.5cm}
    \centering
    \setlength{\tabcolsep}{0.7mm}{
    \begin{tabular}{lcc}
        \toprule
        Model & P@100 & P@10k\\
        \midrule
        PCNN & 72.3 & 81.0\\
        PCNN+ATT & 82.0 & 82.2\\
        BGRU+ATT & 74.0 & 88.1\\
        BGRU+SET & 87.0 & 67.4\\
        PCNN+INTRA+INTER & 91.8& -\\
        ATT+CAPNET & 84.0 & -\\
        QARE+ATT & 93.0 & 91.6\\
        RA-CAPNET & \textbf{98.0} & \textbf{96.6}\\
        \bottomrule
    \end{tabular}}
    \caption{P@100 and P@10k.}
    \label{table:pn}
    \end{minipage}
  \end{table}

A detailed comparison of all approaches, the areas of the PR curves, P@100 and P@10k on NYT-10 or NYT-18, are illustrated in Table~\ref{table:pr} and Table~\ref{table:pn}. From the tables, we find that: (1) RA-CapNet is the first method to increase the PR curve area over 0.5 on NYT-10 while improving it on NYT-18 to 0.7. In P@100 and P@10k, our model also achieves superior performance. This result further demonstrates the effectiveness of RA-CapNet with multi-instance learning on overlapped relation extraction. (2) Capnet-based models achieve better performance on the highly complex NYT-18 dataset, which results from their capability of handling overlapped relations and complex sentences.

\subsection{Ablation Study}

\begin{figure}[htbp]
    \centering
    \subfigure{
    \begin{minipage}[b]{7.5cm}
    \centering
    \includegraphics[scale=0.5]{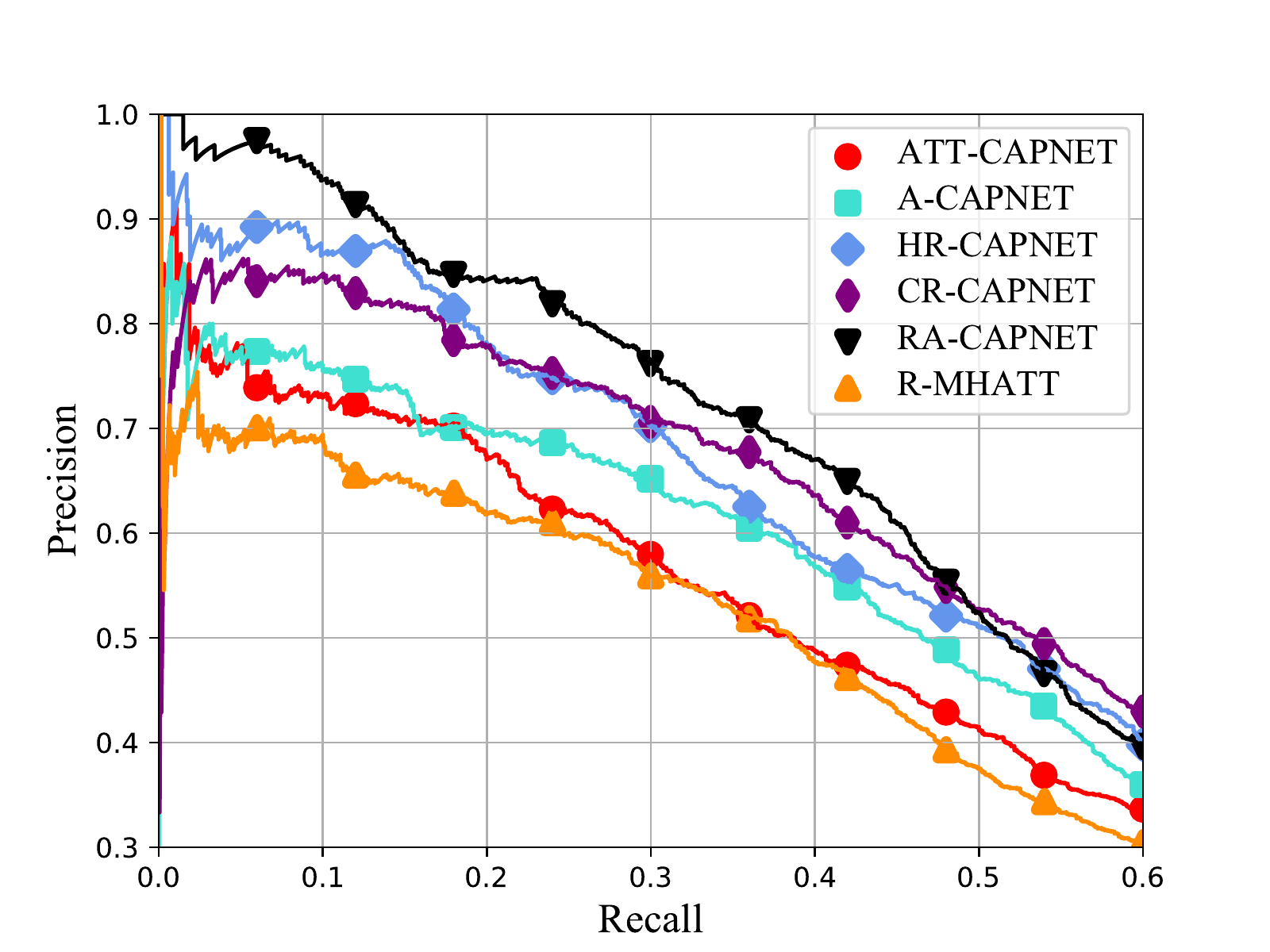}
    \caption{PR curves of our models.}
    \label{fig:ablation-pr}
    \end{minipage}
    }
    \subfigure{
    \begin{minipage}[b]{7.5cm}
        \centering
        \begin{tabular}{lc}
            \toprule
            Model & PR curve area \\
            \midrule
            R-MHATT & 0.383 \\
            ATT-CAPNET  & 0.415 \\
            A-CAPNET  & 0.449 \\
            HR-CAPNET  & 0.493 \\
            CR-CAPNET & 0.501 \\
            RA-CAPNET  & \textbf{0.526} \\
            \bottomrule
        \end{tabular}
        \caption{PR curve areas of our models.}
        \label{table:ablation}
    \end{minipage}
    }
  \end{figure}

To further evaluate the impacts of different parts on RA-CapNet, we compare the performance on the NYT-10 dataset of RA-CapNet with five settings:

\textbf{R-MHATT}: Two multi-head attention layers with relation query attention.

\textbf{ATT-CAPNET}: The same as above. 

\textbf{A-CAPNET}: RA-CapNet without disagreement regularization.

\textbf{HR-CAPNET}: RA-CapNet without capsule disagreement regularization.

\textbf{CR-CAPNET}: RA-CapNet without multi-head disagreement regularization.

\textbf{RA-CAPNET}: Our model.

In Figure~\ref{fig:ablation-pr} and \ref{table:ablation}, the result indicates that: (1) ATT-CAPNET improves the performance of R-MHATT by incorporating the capsule network for handling the multiple relations. (2) Compared with ATT-CAPNET, A-CAPNET improves the PR curve area from 0.415 to 0.449. This proves that relation query multi-head attention helps the capsule network extract salient relation features from different representations at different positions. (3) HR-CAPNET further increases the PR curve area to 0.493, which proves the effectiveness of our disagreement regularization on multiple heads in discriminating the diverse overlapped relation features. (4) Compared with A-CAPNET, CR-CAPNET achieves 0.501 of the PR curve area. This demonstrates that disagreement regularization on the capsules helps models distinguish multiple relation features more clearly. (5) Our complete model, RA-CAPNET, achieves the best performance, showing that the relation query multi-head attention and disagreement regularization term are both effective for relation extraction.

\subsection{Case Study}

\begin{figure}[htbp]
    \centering
    \includegraphics[width=16cm]{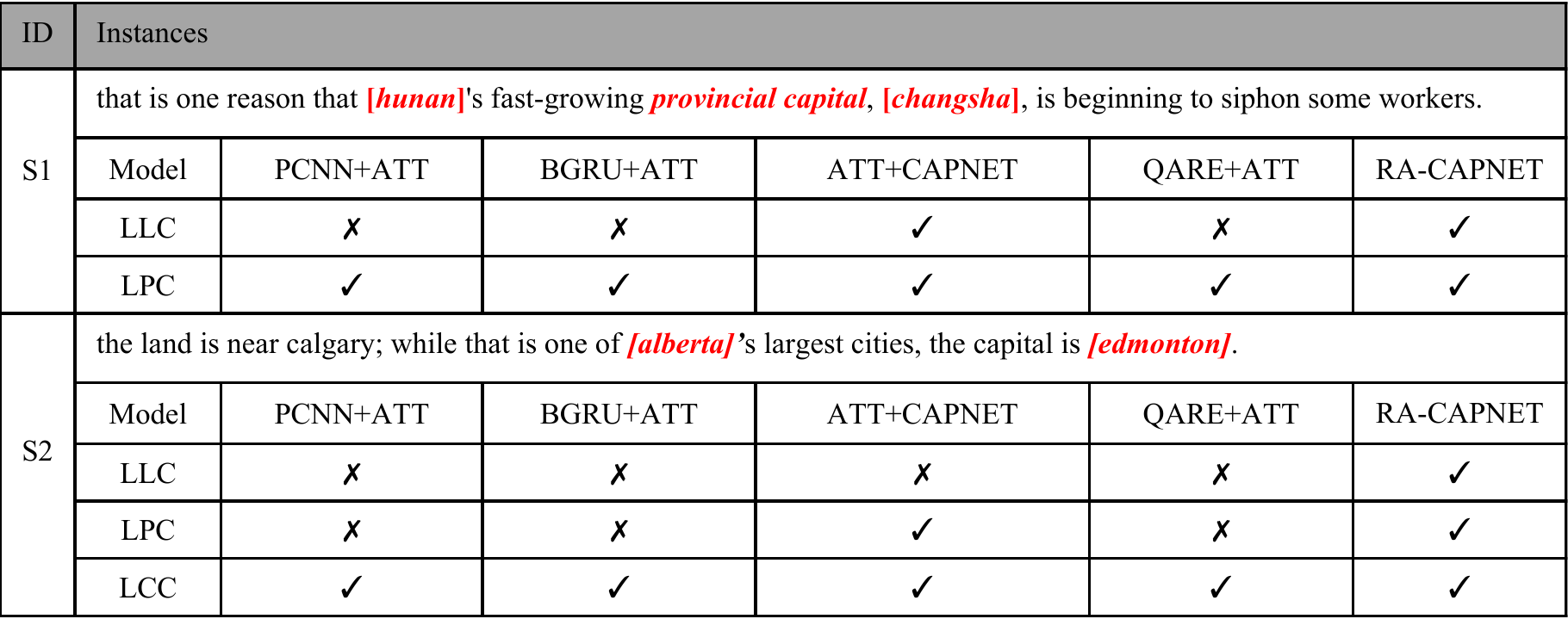}
    \caption{Prediction results of different models on some samples. ``LLC", ``LPC" and ``LCC" represent relation labels of ``/location/location/contains", "/location/province/capital" and "/location/country/capital".}
    \label{fig:case_study}
 \end{figure}

 In Figure~\ref{fig:case_study}, we randomly select two samples from NYT-10 to analyze the prediction performance of different models, where the entities are labeled in the red and bold brackets. From the figure, we find the following: (1) In \textit{S1} and \textit{S2}, compared with CNN/RNN/Attention-based methods, the capsule-based approaches can predict multiple similar relations. (2) In \textit{S2}, only RA-CapNet predicts the correct relation of ``/location/location/contains". This result demonstrates that by incorporating relation query multi-head attention and disagreement regularization in the capsule network, RA-CapNet makes further progress in discriminating overlapped relations.

\section{Conclusion and Future Work}

In this paper, we propose a novel regularized attentive capsule network for overlapped relation extraction. RA-CapNet embeds relation query multi-head attention into the capsule network and uses a novel disagreement regularization term to encourage the diversity among heads and capsules, making it capable of gathering salient information from diverse semantic spaces. Our model is resistant to the noise of distant supervision and achieves significant improvements on both standard and complex datasets.

In the future, we will experiment with different forms of regularization terms and their application to other components of our model.

\section*{Acknowledgements}
This work is partially supported by Chinese National Research Fund (NSFC) Key Project No. 61532013 and No. 61872239; BNU-UIC Institute of Artificial Intelligence and Future Networks funded by Beijing Normal University (Zhuhai) and AI and data Science Hub, United International College (UIC), Zhuhai, Guangdong, China.

\bibliographystyle{coling}
\bibliography{coling2020}

\begin{thebibliography}{}

\bibitem[\protect\citename{Bollacker \bgroup et al.\egroup
  }2008]{bollacker2008freebase}
Kurt Bollacker, Colin Evans, Praveen Paritosh, Tim Sturge, and Jamie Taylor.
\newblock 2008.
\newblock Freebase: a collaboratively created graph database for structuring
  human knowledge.
\newblock In {\em Proceedings of the ACM SIGMOD}.

\bibitem[\protect\citename{Bordes \bgroup et al.\egroup
  }2013]{bordes2013translating}
Antoine Bordes, Nicolas Usunier, Alberto Garcia-Duran, Jason Weston, and Oksana
  Yakhnenko.
\newblock 2013.
\newblock Translating embeddings for modeling multi-relational data.
\newblock In {\em Proceedings of the NeurIPS}.

\bibitem[\protect\citename{Feng \bgroup et al.\egroup
  }2018]{feng2018reinforcement}
Jun Feng, Minlie Huang, Li~Zhao, Yang Yang, and Xiaoyan Zhu.
\newblock 2018.
\newblock Reinforcement learning for relation classification from noisy data.
\newblock In {\em Proceedings of the AAAI}.

\bibitem[\protect\citename{Graves}2013]{graves2013generating}
Alex Graves.
\newblock 2013.
\newblock Generating sequences with recurrent neural networks.
\newblock {\em arXiv preprint arXiv:1308.0850}.

\bibitem[\protect\citename{Hoffmann \bgroup et al.\egroup
  }2011]{hoffmann2011knowledge}
Raphael Hoffmann, Congle Zhang, Xiao Ling, Luke Zettlemoyer, and Daniel~S Weld.
\newblock 2011.
\newblock Knowledge-based weak supervision for information extraction of
  overlapping relations.
\newblock In {\em Proceedings of the ACL}.

\bibitem[\protect\citename{Ji \bgroup et al.\egroup }2017]{ji2017distant}
Guoliang Ji, Kang Liu, Shizhu He, and Jun Zhao.
\newblock 2017.
\newblock Distant supervision for relation extraction with sentence-level
  attention and entity descriptions.
\newblock In {\em Proceedings of the AAAI}.

\bibitem[\protect\citename{Jia \bgroup et al.\egroup }2019]{jia2019arnor}
Wei Jia, Dai Dai, Xinyan Xiao, and Hua Wu.
\newblock 2019.
\newblock Arnor: Attention regularization based noise reduction for distant
  supervision relation classification.
\newblock In {\em Proceedings of the ACL}.

\bibitem[\protect\citename{Kingma and Ba}2014]{kingma2014adam}
Diederik~P Kingma and Jimmy Ba.
\newblock 2014.
\newblock Adam: A method for stochastic optimization.
\newblock {\em arXiv preprint arXiv:1412.6980}.

\bibitem[\protect\citename{Lin \bgroup et al.\egroup }2016]{lin2016neural}
Yankai Lin, Shiqi Shen, Zhiyuan Liu, Huanbo Luan, and Maosong Sun.
\newblock 2016.
\newblock Neural relation extraction with selective attention over instances.
\newblock In {\em Proceedings of the ACL}.

\bibitem[\protect\citename{Liu \bgroup et al.\egroup }2018]{liu2018neural}
Tianyi Liu, Xinsong Zhang, Wanhao Zhou, and Weijia Jia.
\newblock 2018.
\newblock Neural relation extraction via inner-sentence noise reduction and
  transfer learning.
\newblock In {\em Proceedings of the EMNLP}.

\bibitem[\protect\citename{Mikolov \bgroup et al.\egroup
  }2013]{mikolov2013efficient}
Tomas Mikolov, Kai Chen, Greg Corrado, and Jeffrey Dean.
\newblock 2013.
\newblock Efficient estimation of word representations in vector space.

\bibitem[\protect\citename{Mintz \bgroup et al.\egroup }2009]{mintz2009distant}
Mike Mintz, Steven Bills, Rion Snow, and Dan Jurafsky.
\newblock 2009.
\newblock Distant supervision for relation extraction without labeled data.
\newblock In {\em Proceedings of the Joint Conference of the ACL and the
  AFNLP}.

\bibitem[\protect\citename{Miwa and Bansal}2016]{miwa2016end}
Makoto Miwa and Mohit Bansal.
\newblock 2016.
\newblock End-to-end relation extraction using lstms on sequences and tree
  structures.
\newblock In {\em Proceedings of the ACL}.

\bibitem[\protect\citename{Qin \bgroup et al.\egroup }2018a]{qin2018dsgan}
Pengda Qin, XU~Weiran, and William~Yang Wang.
\newblock 2018a.
\newblock Dsgan: Generative adversarial training for distant supervision
  relation extraction.
\newblock In {\em Proceedings of the ACL}.

\bibitem[\protect\citename{Qin \bgroup et al.\egroup }2018b]{qin2018robust}
Pengda Qin, XU~Weiran, and William~Yang Wang.
\newblock 2018b.
\newblock Robust distant supervision relation extraction via deep reinforcement
  learning.
\newblock In {\em Proceedings of the ACL}.

\bibitem[\protect\citename{Riedel \bgroup et al.\egroup
  }2010]{riedel2010modeling}
Sebastian Riedel, Limin Yao, and Andrew McCallum.
\newblock 2010.
\newblock Modeling relations and their mentions without labeled text.
\newblock In {\em Proceedings of the ECML and PKDD}.

\bibitem[\protect\citename{Sabour \bgroup et al.\egroup
  }2017]{sabour2017dynamic}
Sara Sabour, Nicholas Frosst, and Geoffrey~E Hinton.
\newblock 2017.
\newblock Dynamic routing between capsules.
\newblock In {\em Proceedings of the NeurIPS}.

\bibitem[\protect\citename{Surdeanu \bgroup et al.\egroup
  }2012]{surdeanu2012multi}
Mihai Surdeanu, Julie Tibshirani, Ramesh Nallapati, and Christopher~D Manning.
\newblock 2012.
\newblock Multi-instance multi-label learning for relation extraction.
\newblock In {\em Proceedings of the 2012 Joint Conference on EMNLP and CoNLL}.

\bibitem[\protect\citename{Vaswani \bgroup et al.\egroup
  }2017]{vaswani2017attention}
Ashish Vaswani, Noam Shazeer, Niki Parmar, Jakob Uszkoreit, Llion Jones,
  Aidan~N Gomez, {\L}ukasz Kaiser, and Illia Polosukhin.
\newblock 2017.
\newblock Attention is all you need.
\newblock In {\em Proceedings of the NeurIPS}.

\bibitem[\protect\citename{Xu \bgroup et al.\egroup }2015]{xu2015classifying}
Yan Xu, Lili Mou, Ge~Li, Yunchuan Chen, Hao Peng, and Zhi Jin.
\newblock 2015.
\newblock Classifying relations via long short term memory networks along
  shortest dependency paths.
\newblock In {\em Proceedings of the EMNLP}.

\bibitem[\protect\citename{Yang \bgroup et al.\egroup }2018]{yang2018modeling}
Baosong Yang, Zhaopeng Tu, Derek~F Wong, Fandong Meng, Lidia~S Chao, and Tong
  Zhang.
\newblock 2018.
\newblock Modeling localness for self-attention networks.
\newblock In {\em Proceedings of the EMNLP}.

\bibitem[\protect\citename{Ye and Ling}2019]{ye2019distant}
Zhi-Xiu Ye and Zhen-Hua Ling.
\newblock 2019.
\newblock Distant supervision relation extraction with intra-bag and inter-bag
  attentions.
\newblock In {\em Proceedings of the ACL}.

\bibitem[\protect\citename{Zeng \bgroup et al.\egroup }2014]{zeng2014relation}
Daojian Zeng, Kang Liu, Siwei Lai, Guangyou Zhou, and Jun Zhao.
\newblock 2014.
\newblock Relation classification via convolutional deep neural network.
\newblock In {\em Proceedings of COLING}.

\bibitem[\protect\citename{Zeng \bgroup et al.\egroup }2015]{zeng2015distant}
Daojian Zeng, Kang Liu, Yubo Chen, and Jun Zhao.
\newblock 2015.
\newblock Distant supervision for relation extraction via piecewise
  convolutional neural networks.
\newblock In {\em Proceedings of the EMNLP}.

\bibitem[\protect\citename{Zhang \bgroup et al.\egroup }2019]{zhang2019multi}
Xinsong Zhang, Pengshuai Li, Weijia Jia, and Hai Zhao.
\newblock 2019.
\newblock Multi-labeled relation extraction with attentive capsule network.
\newblock In {\em Proceedings of the AAAI}.

\bibitem[\protect\citename{{Zhang} \bgroup et al.\egroup }2020]{8952645}
X.~{Zhang}, T.~{Liu}, P.~{Li}, W.~{Jia}, and H.~{Zhao}.
\newblock 2020.
\newblock Robust neural relation extraction via multi-granularity noises
  reduction.
\newblock {\em IEEE Transactions on Knowledge and Data Engineering}.

\bibitem[\protect\citename{Zhou \bgroup et al.\egroup }2016]{zhou2016attention}
Peng Zhou, Wei Shi, Jun Tian, Zhenyu Qi, Bingchen Li, Hongwei Hao, and Bo~Xu.
\newblock 2016.
\newblock Attention-based bidirectional long short-term memory networks for
  relation classification.
\newblock In {\em Proceedings of the ACL}.

\end{thebibliography}

\end{document}